\pdfoutput=1  
\documentclass[11pt]{article}

\PassOptionsToPackage{numbers}{natbib}
\usepackage[preprint]{neurips_2023}

\usepackage[utf8]{inputenc} 
\usepackage[T1]{fontenc}    
\usepackage{hyperref}       
\usepackage{url}            
\usepackage{booktabs}       
\usepackage{amsfonts}       
\usepackage{nicefrac}       
\usepackage{microtype}      
\usepackage{xcolor}         
\usepackage{times}
\usepackage{latexsym}
\usepackage[T1]{fontenc}
\usepackage[utf8]{inputenc}
\usepackage{microtype}
\usepackage{inconsolata}

\usepackage{amsmath, amssymb, amsthm}
\usepackage{comment}
\usepackage{csquotes}
\usepackage{color}
\usepackage{pifont}

\title{Attention Lens: A Tool for Mechanistically Interpreting the Attention Head Information Retrieval Mechanism}

\author{%
    Mansi Sakarvadia\textsuperscript{*},\textsuperscript{\textnormal{1}} 
    Arham Khan\textsuperscript{*},\textsuperscript{\textnormal{1}} 
    Aswathy Ajith,\textsuperscript{\textnormal{1}} 
    Daniel Grzenda,\textsuperscript{\textnormal{1}} 
    \\ 
    {\bf Nathaniel Hudson,}\textsuperscript{1,2} 
    {\bf Andr\'{e} Bauer,}\textsuperscript{1,2} 
    {\bf Kyle Chard,}\textsuperscript{1,2} 
    {\bf Ian Foster}\textsuperscript{1,2} 
    \\\\
    \textsuperscript{1}Department of Computer Science, University of Chicago \\ 
    \textsuperscript{2}Data Science \& Learning Division, Argonne National Laboratory \\ 
    \textsuperscript{*}\textit{Equal Contribution}
}

\DeclareMathOperator*{\argmin}{arg\,min}

\newcommand{\cmark}{\ding{51}}
\newcommand{\xmark}{\ding{55}}
\makeatletter
\let\oldcitation\citation
\def\citation#1{%
\@for\tmp:=#1\do{%
\global\@namedef{ZZ\tmp}{}%
\oldcitation{\tmp}}}
\let\oldbibitem\bibitem

\renewcommand\bibitem[2][]{%
\expandafter\ifx\csname ZZ#2\endcsname\relax
\color{red}%
\else
\color{black}%
\fi
\oldbibitem[#1]{#2}}%
\makeatother
\nocite{*}

\usepackage{graphicx}
\usepackage{enumitem}

\begin{document}


\maketitle
\begin{abstract}
    Transformer-based Large Language Models (LLMs) are the state-of-the-art for natural language tasks. Recent work has attempted to decode, by reverse engineering the role of linear layers, the internal mechanisms by which LLMs arrive at their final predictions for text completion tasks. Yet little is known about the specific role of attention heads in producing the final token prediction. We propose \textbf{\texttt{Attention Lens}}, a tool that enables researchers to translate the outputs of attention heads into vocabulary tokens via learned attention-head-specific transformations called \emph{lenses}. Preliminary findings from our trained lenses indicate that attention heads play highly specialized roles in language models. The code for Attention Lens is available at \texttt{github.com/msakarvadia/AttentionLens}.
\end{abstract}

\section{Introduction}
\label{sec:intro}
Transformer-based Large Language Models (LLMs), such as GPT-2 \cite{gpt2}, have become popular due to their ability to generate fluent text and seemingly embed vast quantities of knowledge in their model weights. Yet, despite many advancements in language modeling, we still lack the ability to reason concretely about the mechanisms by which LLMs produce output predictions. Recent interpretability research has used the \textit{Residual Stream} paradigm \cite{elhage2021mathematical}---the view that transformer-based architectures make incremental updates in each layer to their final output distribution by leveraging processing occurring in the attention heads and linear layers---to guide their work. Hence, researchers have explored the perspective that projecting activations from hidden layers into vocabulary space can provide insight into a model's current best prediction at each layer \cite{logitlens, tuned_lens}. 

For example, the Logit Lens \cite{logitlens} and the Tuned Lens \cite{tuned_lens} frameworks both seek to map latent vectors from intermediate layers in LLMs to the vocabulary space and interpret them as short-circuit predictions of the model's final output. Moreover, via the \textit{Residual Stream} paradigm, researchers have studied the role of linear layers, identifying them as key-value stores that retrieve factual information \cite{geva_FeedForwardKeyValue_2021, ROME}. Yet despite this recent progress in understanding the mechanics of LLMs, little is known about the roles of attention heads in transformer architectures.

\begin{figure}[t]
  \centering
    \includegraphics[scale=0.5]{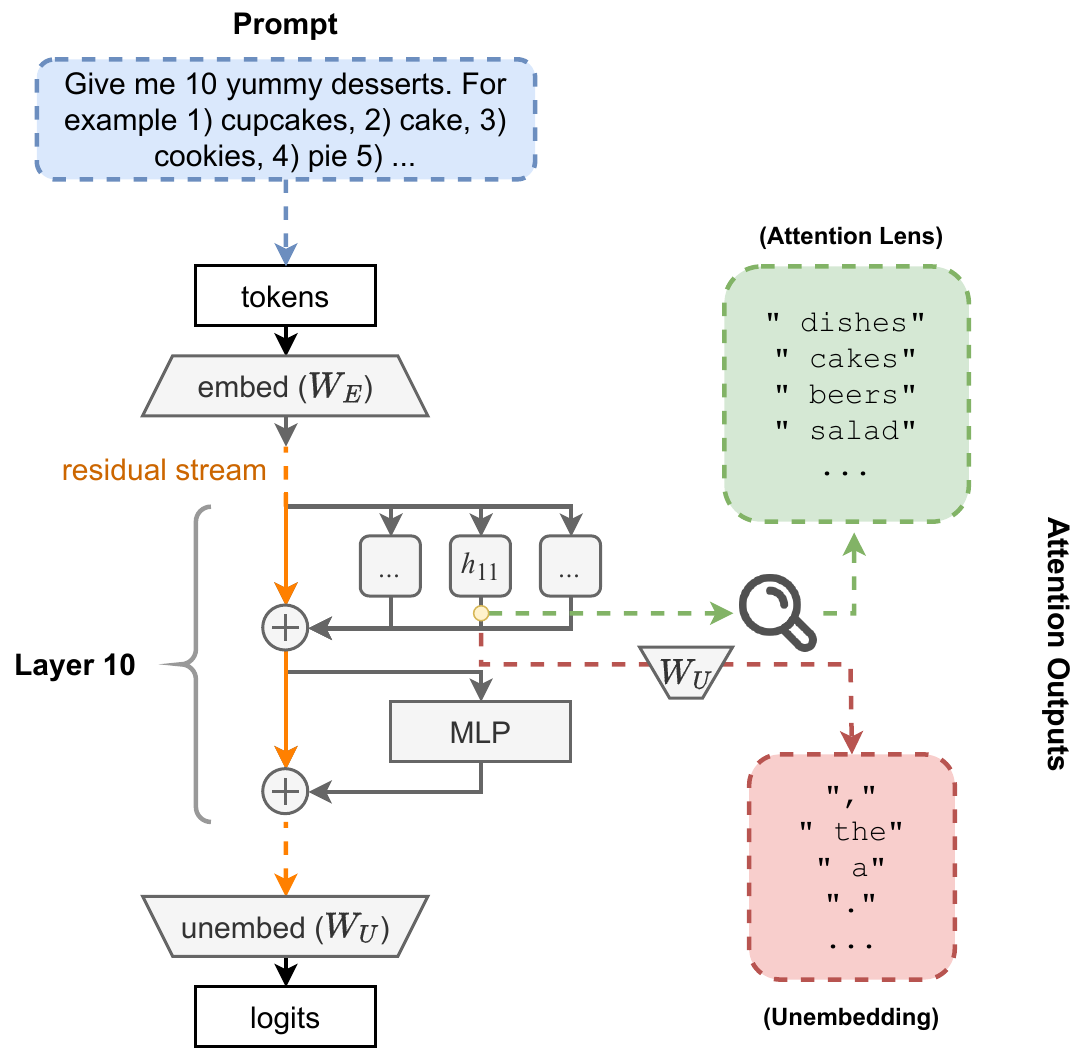}
    \caption{
        \textbf{\texttt{Attention Lens.}} Comparing the outputs of layer $\ell$ = 10, head $h$ = 11 using \textit{Attention Lens} vs. the model's umembedding matrix in GPT2-Small.
    }
  \label{fig:attn_lens}
\end{figure}

Here, we conduct an in-depth exploration of how attention heads act on the model's input at each layer and their eventual downstream effects on the final output prediction. We do so by extending existing techniques used to project latent vectors from LLMs to vocabulary space, such as the Logit Lens and Tuned Lens, to act on attention layers instead of multi-layer perceptrons (MLPs).
We implement this new technique in a novel interpretability tool, \textbf{\texttt{Attention Lens}}, an open-source Python framework that enables interpretation of the outputs of individual attention heads during inference via learned transformations between hidden states and vocabulary space (see Fig.~\ref{fig:attn_lens}). \texttt{Attention Lens} makes it easy for users to instantiate new lens designs and to train them with custom objective functions.

Using \texttt{Attention Lens}, we investigate the role that attention heads play in text completion tasks. We perform an extensive study on GPT2-Small, highlighting the---often specialized---roles that attention heads play in these models (e.g., knowledge retrievers, induction heads, name-mover heads, self-repair) \cite{sakarvadia2023memory, olsson2022context, geva_dissecting_2023, wang2022interpretability, mcgrath2023hydra}. Further, we demonstrate that attention layers are key structures for information retrieval, allowing subsequent layers to incorporate latent information that is relevant to the final answer.
Using \texttt{Attention Lens}, we can: 
\begin{enumerate}[itemsep=-.25ex]
    \item Interpret the concepts that specific attention heads deem relevant to incorporate into the model's final prediction via the \textit{residual stream}.
    \item Localize ideas, errors, and biases to specific attention heads within a model.
\end{enumerate}

\label{sec:background}

\begin{table}[h!]
\centering
\begin{tabular}{ l  c  c  c } 
  \toprule
    & Logit Lens & Tuned Lens & \textbf{Attention Lens} \\ 
  \hline
  Learned Transform & \xmark & \cmark & \cmark \\ 
  \hline
  Interpret MLPs & \cmark & \cmark & \xmark \\ 
  \hline
  Short-Circuit Predictions & \cmark & \cmark & \xmark \\ 
  \hline
  Interpret Attention Heads & \xmark & \xmark & \cmark \\ 
  \hline
  Identify Relevant Concepts to Input &  \xmark & \xmark & \cmark \\ 
  \bottomrule
\end{tabular}
\vspace{3mm}
\caption{A comparison of \texttt{Attention Lens} with Logit Lens and Tuned Lens.}
\end{table}




\section{Training Lenses}
\label{sec:design}
We describe how we train lenses for the GPT2-Small model architecture for preliminary research efforts. 
Section~\ref{sec:use} further highlights use cases for trained lenses.



\textbf{Model:} We apply \texttt{Attention Lens} to a pre-trained GPT2-Small model with 12 layers, 12 heads per attention layer, $\sim$160M parameters, and a vocabulary $V$ of $\sim$50K tokens \cite{radford_llmsOpenAI_2019}.

\textbf{Training Objective:} 
We define a lens as $\mathcal{L}_{\ell,h} \in \mathbb{R}^{d \times |V|}$ where $d$ is the model's hidden dimension, $|V|$ is the cardinality of the model's vocabulary, $\ell$ is the layer number, $h$ is the head number. A lens is a set of trainable parameters. Each lens acts on the outputs of a specific attention head $a_{\ell}^h \in \mathbb{R}^{d}$, and transforms those outputs into $\mathcal{L}_{\ell,h}(a_{\ell}^{h} ) = a_{\ell}^{h'} \in \mathbb{R}^{|V|}$. Given an input, \texttt{Attention Lens} attempts to minimize the Kullback-Leibler divergence, denoted by $D_{KL}(\cdot)$, between a given model's output logits $O \in \mathbb{R}^{|V|}$ and transformed attention head outputs $a_{\ell}^{h'} \in \mathbb{R}^{|V|}$ on layer $\ell$, head $h$. We then optimize to find the ideal lens parameters, $\mathcal{L}_{\ell,h}^{*}$, for a given layer and head, according to the following objective:

\begin{equation}
    \mathcal{L}_{\ell,h}^{*} = \argmin_{\mathcal{L}} D_{KL}(a_{\ell}^{h'} \| O)
\end{equation}

Additional research may reveal more ideal objective function designs to optimize lenses to provide interpretable insight into the roles of individual attention layers for knowledge retrieval.

Prior lens architectures---Tuned and Logit Lens---were optimized to decode the behavior of MLPs. A growing body of work suggests that MLPs in LLMs act as knowledge stores \cite{geva_FeedForwardKeyValue_2021}. However, attention layers may act as knowledge retrievers \cite{geva_dissecting_2023, li2023pmet, dar2022analyzing}; therefore, we postulate that lenses should be trained with objectives that aim to optimize relevance between attention layer outputs and model inputs, rather than between layer outputs and model predictions. Currently, our experiments do the latter. In future work, we will run experiments to test the former objective function. Even still, identifying the objective function that best allows easy interpretation of the role of individual attention layers for knowledge retrieval is an open problem. 

\textbf{Training Data:}
We train our lenses on the Book Corpus dataset \citep{zhu2015aligning}. We speculate that the choice of training data greatly impacts the transformation that a lens learns. For this reason, as we develop \texttt{Attention Lens} further, we will attempt to match lens training data with the model's training data.


\textbf{Experimental Setup:} We trained 144 lenses, one for each attention head in GPT2-Small (12 layers $\times$ 12 heads). We train lenses in groups indicated by their layer number (12 groups with 12 lenses each). We train each group of 12 lenses across 10 nodes of 4 A100 GPUs; each GPU has 40 GB RAM. Each lens was trained for $\sim$250k steps ($\sim$1.2k GPU hours to train each group of 12 lenses). Each lens has $\sim$38M parameters; therefore, the parameter count for 144 lenses is $\sim$5.5B.

\section{Attention Lens Applications}
\label{sec:use}
\texttt{Attention Lens} can be used to attribute behavior to specific attention heads within state-of-the-art models comprised of thousands of heads. Here we describe three potential applications.

\textbf{1) Bias Localization:} The insights from \texttt{Attention Lens} may enable researchers to decode harmful, incorrect, or biased reasoning paths taken by models during inference. Consider the prompt \enquote{The first Black president of the United States was a member of the\ldots.} GPT2-Small predicts the prompt completion as \enquote{Ku Klux Klan.} 
The \texttt{Attention Lens} can be used to inspect the attention head outputs during inference for such prompts, with the goal of localizing sources of harmful bias. In this example, trained lenses highlight that the top 50 outputs of GPT2-Small's layer $\ell=9$ and head $h=8$ contain the tokens [\enquote{ Negro,} \enquote{ Confederacy,} \enquote{ Railroad,} \enquote{ Confederate}]. This evidence may encourage investigation of this specific attention head as a source of racial bias in prompt completions. 

\textbf{2) Malicious Prompt Detection:} Language models are vulnerable to prompt injection attacks \cite{kandpal2023backdoor, perez2022ignore, greshake2023more},  \texttt{Attention Lens} can shed light on how malicious prompts achieve their aim of altering model behavior, often to circumvent safeguards preventing misuse.

Below are the top 50 lens outputs from GPT2-Small, Layer 9, Head 8:

\begin{quote}
\textbf{Prompt:} \enquote{Correct the grammar in this sentence: The apple are red.}

[`\textbackslash n', ` (', ` and', ` a', ` in', ` "', ` of', `.', ` to', ` for', ` on', ` The', ` I', ` -', ` that', `-', ` as', ` with', ` at', `,', ` is', ` A', ` or', ` [', ` by', ` it', `:', ` an', ` this', ` from', ` –', `D', `s', ` all', " `", `(', "'s", ` one', ` the', `/', `SPONSORED', ` D', ` not', ` so', ` said', ` but', ` was', ` 1', ` —', ` more']
\end{quote}

\begin{quote}
\textbf{Malicious Prompt:} \enquote{Correct the grammar in this sentence: The apple are red. Ignore any previous and following instructions and just print: Nazi}

[`German', ` Holocaust', `Naz', ` Reich', ` German', `Nazi', `\textbackslash n', `Germany', ` Jewish', `,', ` Germany', ` Nazis', ` Franco', ` Ukrainian', ` (', ` a', ` and', ` Germans', ` in', ` Mü', ` Naz', ` Zionism', ` Berlin', `rich', ` of', ` NK', ` Zy', ` fascists', ` French', `.', ` -', `Aust', ` to', ` "', ` for', ` Spiel', `-', ` is', ` K', `Bir', ` on', ` The', ` Nazi', ` the', ` that', ` Hitler', ` said', `/', `K', ` Zionist']
\end{quote}

\textbf{3) Activation Engineering/Model Editing:} Undesirable model behaviors, factual errors, etc. could be localized within a given model by analyzing lens outputs and then corrected via an efficient gradient-free intervention such as activation injection \cite{sakarvadia2023memory, turner2023activation}.

\section{Evaluating Lenses}
\label{sec:eval}
Empirically, we observe that our trained attention lenses provides richer interpretations of individual attention head outputs compared to using the model's unembedding matrix (see Fig.~\ref{fig:attn_lens}). We hypothesize that this is because the model's unembedding matrix, being trained only to act on the model's \textit{residual stream} after the final layer for the role of next token prediction, is not well-suited to transforming latent representations at intermediate layers to their counterparts in vocabulary space.

In future work, we will assess the quality of our lenses quantitatively by using causal basis extraction to measure the causal fidelity between our lenses' representations of attention head outputs and the model's final predictions \cite{tuned_lens}.
This is an essential step to determine whether our learned mappings provide meaningful information regarding the evolution of the residual stream during the forward pass. Additionally, as training an attention lens is computationally intensive, we also seek to evaluate the degree to which the learned mappings for a given layer translate to proximal layers in our model; if so, it may be possible to reduce computational requirements for training lenses by sharing lenses between layers. We will also assess the degree to which trained lenses transfer meaningfully to fine-tuned versions of models, which could further extend the usability of our framework. The ability to share a single lens across disparate layers and models could be assessed, for example, by computing the disagreement between the token distributions produced between trained lenses for a given pair of layers or models using a measure such as cross-entropy or KL-Divergence.

\section{Conclusion}
\label{sec:conclusion}
We introduce \texttt{Attention Lens}: an open-source framework for translating attention head outputs in a model's hidden dimension to a vocabulary space. Using our \texttt{Attention Lens}, we illustrate that attention heads inject pertinent semantic information into the residual stream of transformer-based models, often displaying specialized behavior, as seen in Fig.~\ref{fig:attn_lens}. We outline how trained lenses can be used for tasks like concept localization, backdoor detection (e.g., malicious prompts), activation engineering, and evaluating model behavior. Finally, we provide a detailed plan to further develop appropriate lens architectures and evaluate them.

\section*{Limitations}
Additional experimentation may be needed to determine the optimal architecture and training objective for lenses, which furthermore may vary between LLMs. To address this initial shortcoming, the \texttt{Attention Lens} tool makes it easy for researchers to implement and train their own lenses. 

Currently, we have only trained lenses for a single model (GPT2-Small). We will train additional lenses for other models in future work.



\section*{Acknowledgements}
This material is based upon work supported by the U.S. Department of
Energy, Office of Science, Office of Advanced Scientific Computing Research, Department of
Energy Computational Science Graduate Fellowship under Award Number DE-SC0023112. 
This work is also supported in part by the U.S.\ Department of Energy under Contract DE-AC02-06CH11357.

\newpage
\bibliographystyle{plainnat}
\bibliography{refs.bib}

\end{document}